\documentclass[acmsmall,screen,nonacm]{acmart}
\usepackage{url}
\usepackage{microtype}
\usepackage{graphicx}
\usepackage{subcaption}
\usepackage{multirow}

\usepackage{ulem} 
\usepackage{xcolor}
\newcommand\hl{\bgroup\markoverwith
  {\textcolor{yellow}{\rule[-.5ex]{.1pt}{2.5ex}}}\ULon}

\begin{document}

\title{Explore as a Storm, Exploit as a Raindrop: On the Benefit of Fine-Tuning Kernel Schedulers with Coordinate Descent}

\author{Michael Canesche}
  \email{michaelcanesche@dcc.ufmg.br}
  \orcid{0000-0001-7882-0787}
\affiliation{
  \institution{UFMG}
  \country{Brazil}
}
\email{michaelcanesche@dcc.ufmg.br}

\author{Gaurav Verma}
  \email{gaurav.verma@stonybrook.edu}
  \orcid{0000-0002-8820-6016}
\affiliation{
  \institution{Stony Brook University}
  \country{USA}
}
\email{gaurav.verma@stonybrook.edu}          

\author{Fernando Magno Quint\~{a}o Pereira}
  \email{fernando@dcc.ufmg.br}
  \orcid{0000-0002-0375-1657}
\affiliation{
  \institution{UFMG}             
  \country{Brazil}
}
\email{fernando@dcc.ufmg.br}          

\begin{abstract}
Machine-learning models consist of kernels, which are algorithms applying operations on tensors---data indexed by a linear combination of natural numbers. Examples of kernels include convolutions, transpositions, and vectorial products. There are many ways to implement a kernel. These implementations form the kernel's optimization space. Kernel scheduling is the problem of finding the best implementation, given an objective function---typically execution speed. Kernel optimizers such as \texttt{Ansor}, \texttt{Halide}, and \texttt{AutoTVM} solve this problem via search heuristics, which combine two phases: exploration and exploitation. The first step evaluates many different kernel optimization spaces. The latter tries to improve the best implementations by investigating a kernel within the same space. For example, \texttt{Ansor} combines kernel generation through sketches for exploration and leverages an evolutionary algorithm to exploit the best sketches. In this work, we demonstrate the potential to reduce \texttt{Ansor}'s search time while enhancing kernel quality by incorporating Droplet Search, an \texttt{AutoTVM} algorithm, into \texttt{Ansor}'s exploration phase. The approach involves limiting the number of samples explored by \texttt{Ansor}, selecting the best, and exploiting it with a coordinate descent algorithm. By applying this approach to the first 300 kernels that \texttt{Ansor} generates, we usually obtain better kernels in less time than if we let \texttt{Ansor} analyze 10,000 kernels. This result has been replicated in 20 well-known deep-learning models (AlexNet, ResNet, VGG, DenseNet, etc.) running on four architectures: an AMD Ryzen 7 (x86), an NVIDIA A100 tensor core, an NVIDIA RTX 3080 GPU, and an ARM A64FX. A patch with this combined approach was approved in \texttt{Ansor} in February 2024. As evidence of the generality of this search methodology, a similar patch, achieving equally good results, was submitted to \texttt{TVM}'s \texttt{MetaSchedule} in June 2024.
\end{abstract}




\maketitle


\section{Introduction}
\label{sec:introduction}

A \textit{kernel} is an algorithm that applies operations on \textit{tensors}: chunks of memory indexed by a linear combination of natural numbers.
A \textit{Tensor Compiler} is a compilation infrastructure that generates code for kernels.
As \citet{Ansel24} explains, many tensor compilers, including \texttt{TVM}~\cite{Chen18}, \texttt{nvFuser}~\cite{Sarofeen22} and \texttt{NNC}~\cite{Sarofeen21}, follow a design probably inspired by \texttt{Halide}~\cite{Kelley13}.
These compilers separate the kernel's semantics (what the kernel does) from its schedule (when the kernel does it).
Since the same kernel semantics can be implemented in many different schedules, tensor compilers face a challenge called \textit{kernel scheduling}: determining a suitable ordering for the operations a kernel performs on tensors.
Kernel scheduling is typically addressed via heuristics because the \textit{Kernel Optimization Space}---the set of all implementations of a kernel---is extremely vast, as \citet{Li21} explains.

The \texttt{Apache TVM} tensor compiler employs three distinct optimization infrastructures for solving kernel scheduling: \texttt{AutoTVM}~\cite{Chen18},
\texttt{Ansor}~\cite{Zheng20} and \texttt{MetaSchedule}~\cite{Shao21}.
\texttt{AutoTVM} finds parameters of \textit{kernel sketches}.
The sketch of a kernel represents the optimizations applied to the abstract description of that kernel, such as loop unrolling, splitting, interchange, and tiling.
Many of these optimizations are parameterizable.
Examples of parameters include the unrolling factor in loop unrolling and the width of the tiling window in loop tiling.
\texttt{AutoTVM} assigns values to these parameters using various search heuristics.
One of these heuristics is of interest in this paper: Droplet Search~\cite{Canesche24}.
Droplet Search seeks the optimal configuration of an optimization template by determining a descent direction along the objective function that models the running time of the kernel.
\texttt{Ansor} and \texttt{MetaSchedule} (in contrast to \texttt{AutoTVM}) have the capability to generate new sketches.
In other words, these schedulers are not restricted to a single sequence of optimizations.
Section~\ref{sub:ansor_gen} provides further details on how \texttt{Ansor} works, whereas Section~\ref{sub:droplet_exp} explains how Droplet Search works.

\paragraph{A Combined Search Infrastructure.}
Both \texttt{Ansor} and \texttt{MetaSchedule} typically generate higher-quality kernels compared to \texttt{AutoTVM}'s Droplet Search, as they are not limited to a single search space.
Each new sketch leads to the exploration of an entirely new search space.
However, when constrained to a single sketch, Droplet Search tends to outperform these schedulers. 
Drawing on the terminology from recent work by \citet{Ding23}, Droplet Search's coordinate descent approach is ``{\it hardware centric}'', while \texttt{Ansor}'s genetic algorithm is ``{\it input centric}''; better embodying this quality that \citet{Sorensen19} calls ``{\it performance portability}''.
In essence, Droplet Search, by traversing the search space contiguously, is sensitive to cache sizes and levels in the cache hierarchy.
Nevertheless, despite Droplet Search's tendency to identify optimal points within the search space, this algorithm is unable to navigate beyond this space due to its reliance on the initial sketch.

Inspired by these observations, this paper addresses the following research question: ``{\it Is it possible to combine the wide exploration phase of \texttt{Ansor}\footnote{This paper uses \texttt{Ansor} as the basis for experiments; however, similar results have been reproduced in the \texttt{MetaSchedule}~\cite{Canesche24MS}. We focus on \texttt{Ansor} because it is documented in an academic work~\cite{Zheng20}.} with Droplet Search's exploitation; thus, obtaining the advantages of each approach?}'' By doing so, we can develop a version of \texttt{Ansor} that produces superior kernels compared to the original tool while also reducing search times.
This paper brings evidence that such a combination is effective.
The core idea presented in this work is as follows: Initially, we allow \texttt{Ansor} to explore the kernel optimization space, leveraging its ``space travel ability'' to test different sequences of optimizations during this exploration.
Subsequently, following this initial exploration phase, we identify the most promising kernel space discovered by \texttt{Ansor} and employ \texttt{AutoTVM}'s Droplet Search---a line search algorithm---to find a good kernel within this space.

\paragraph{Summary of Findings.}
This paper describes findings of an eminently empirical nature.
The search techniques discussed in Section~\ref{sec:implementation} are not an original contribution of this work: \texttt{Ansor}'s exploration algorithm was designed by \citet{Zheng20}, and Droplet Search was incorporated into \texttt{AutoTVM} by \citet{Canesche24}, and further explored by \citet{Chendi24}.
Nevertheless, combining these two techniques into a practical tool required a number of experimental observations and engineering decisions, which Section~\ref{sec:results} organizes into six research questions.

The new search methodology that emerged out of this combination has been considered sufficiently practical to be approved into \texttt{Apache TVM}.
A request for comments was submitted to the \texttt{TVM} community in December of 2023, and a patch was approved into \texttt{Ansor} in February of 2024.
The patch has not yet been merged into an official release of \texttt{TVM} at the time of this submission.
We have, subsequently, incorporated a coordinate descent exploitation phase into \texttt{MetaSchedule}~\cite{Shao21}, which uses different search algorithms than \texttt{Ansor}.
These results are even better than those observed in \texttt{Ansor}.
Consequently, a new request for comments was submitted to the TVM community in June 2024, together with a patch for \texttt{MetaSchedule}~\cite{Canesche24MS}.

The experiments in Section~\ref{sec:results} show that the combined exploration-exploitation methodology outperforms the original implementation of \texttt{Ansor} in terms of kernel quality and search speed.
Positive results are reported in four different processors (AMD R7-3700X, Fujitsu ARM A64FX, NVIDIA RTX 3080, and NVIDIA A100), and in 20 popular deep-learning models, including \texttt{AlexNet}, \texttt{VGG}, \texttt{ResNet}, \texttt{MobileNet}, \texttt{Inception}, \texttt{GoogleNet}, and \texttt{DenseNet}.
As an illustration, by terminating \texttt{Ansor} after sampling 300 kernels and then optimizing the best candidate with Droplet Search, we outperformed \texttt{Ansor} running with a budget of 10,000 samples in all the $4 \times 20$ architecture-model pairs.
For instance, the combined search approach, on  \texttt{MnasNet}, yields kernel speedups of 1.59x, 1.02x, and 1.08x on x86, ARM, and NVIDIA platforms, respectively. The search time for the combined approach correspondingly decreases by 1.25x, 1.25x, and 1.18x on these architectures.

\section{Exploration via \texttt{Ansor}; Exploitation via Droplet Search}
\label{sec:implementation}

A kernel is an abstract concept: it can be represented by operations on memory indexed by a linear combination of natural numbers.
The actual implementation of a kernel is determined by its {\it schedule}.
The schedule of a kernel determines in which order the different memory elements are accessed when the kernel runs.
Following the terminology introduced in the original \texttt{Ansor} work~\cite{Zheng20}, a schedule is the combination of two notions: a {\it sketch} and an {\it annotation} of the sketch.
Definition~\ref{def:space} enumerates these notions.

\begin{definition}[The Kernel Search Space]
\label{def:space}
The na\"{i}ve implementation of a kernel replaces each linear index in the abstract representation of the kernel with a loop.
A sketch is a sequence of transformations, such as loop fusion, splitting, or tiling, that can be applied to the na\"{i}ve implementation of the kernel.
An annotation of the sketch is the set of parameters that control the effect of each optimization in the sketch, such as unrolling factor, length of tiling window, number of threads in parallelization, etc.
We call an annotated sketch a ``kernel''.
A kernel is a concrete program: it effectively runs.
Each sketch determines a kernel search space, which is the set of every valid way to annotate that sketch.
\end{definition}

\begin{figure}[ht]
\centering
\includegraphics[width=\textwidth]{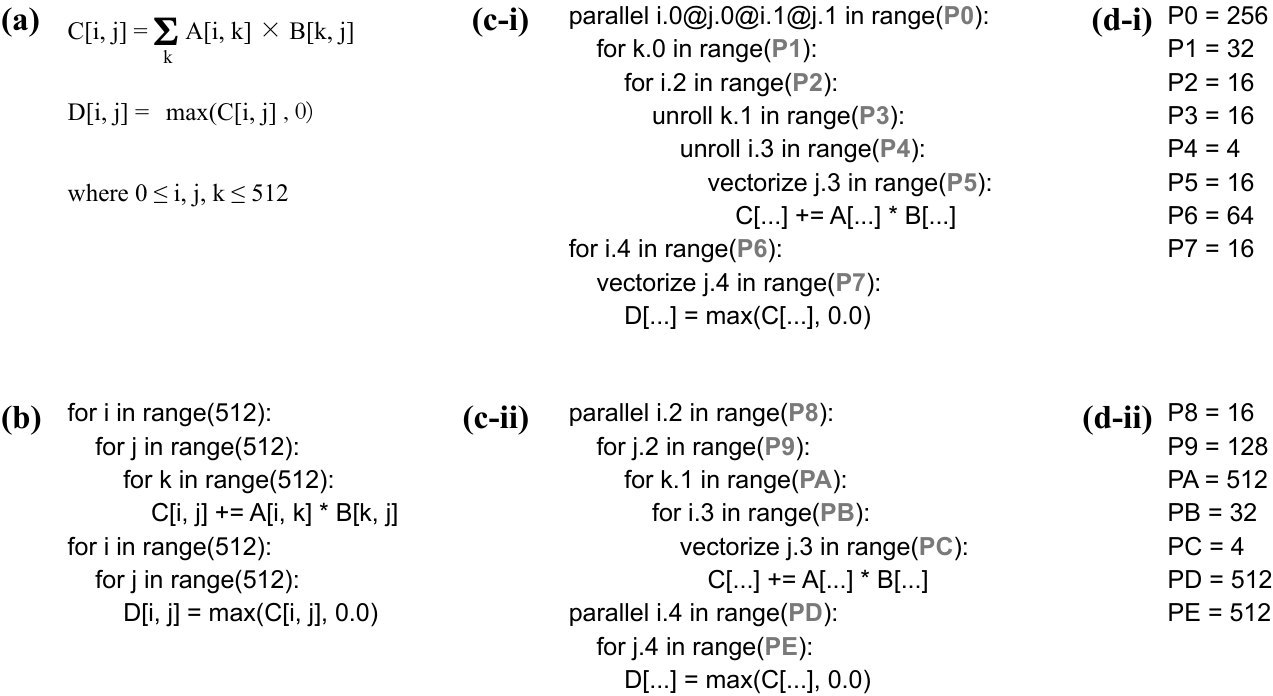}
\caption{(a) Abstract view of a kernel.
(b) Na\"{i}ve implementation of the abstract kernel.
(c) Two optimization sketches for the na\"{i}ve kernel.
(d) Different annotations for the sketches.}
\label{fig:abstract_kernel}
\end{figure}

\begin{example}
\label{ex:sketches}
Figure~\ref{fig:abstract_kernel} (a) shows an example of an abstract kernel.
Figure~\ref{fig:abstract_kernel} (b) shows a na\"{i}ve implementation of the abstract kernel
seen in Figure~\ref{fig:abstract_kernel} (a).
Figure~\ref{fig:abstract_kernel} (c) shows two sketches produced after the application of different code optimizations onto the program in Figure~\ref{fig:abstract_kernel} (b).
\end{example}

\begin{figure}[ht]
\centering
\includegraphics[width=\textwidth]{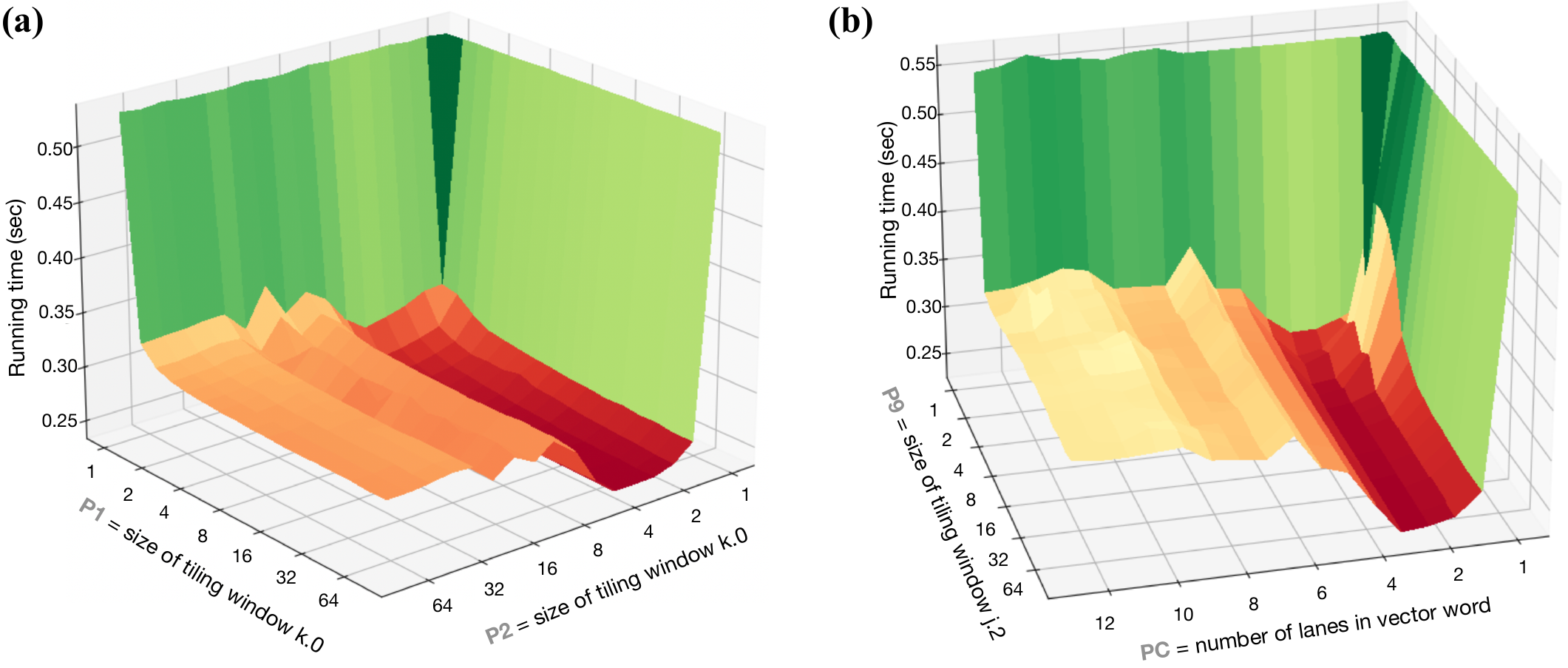}
\caption{(a) A three-dimensional view of the optimization space formed by the parameters \textbf{P1} and
\textbf{P2} seen in Figure~\ref{fig:abstract_kernel} c-i.
(b) A three-dimensional view of the optimization space
of parameters \textbf{P9} and \textbf{PC}.}
\label{fig:kernel_space}
\end{figure}

Figure~\ref{fig:abstract_kernel} (d) shows different parameters of the sketches in Figure~\ref{fig:abstract_kernel} (c).
As introduced in Definition~\ref{def:space}, the set of every valid configuration of annotations for a given sketch forms the {\it search space} of that sketch.
This space has one dimension for each parameter that is allowed to vary.
Figure~\ref{fig:kernel_space} shows two views of the optimization space of the two sketches in Figure~\ref{fig:abstract_kernel} (c).

\subsection{Space Exploration via \texttt{Ansor}}
\label{sub:ansor_gen}

\texttt{Ansor} solves kernel scheduling in an interactive process that involves three phases:
\begin{description}
    \item [Sketch generation:] in this phase, new sketches are created.
    As explained in Definition~\ref{def:space}, each sketch determines one kernel search space.
    \item [Sketch annotation:] in this phase, an initial population of annotated sketches is created.
    Following Definition~\ref{def:space}, each annotated kernel is a point in the search space determined by the sketch that provides the annotations.
    \item [Kernel evolution:] in this phase, candidate kernels are sorted according to an estimation of their performance (via \texttt{Ansor}'s cost model).
    The best candidates are sampled (executed and timed), and this information is used to improve the cost model.
\end{description}

\begin{figure}[ht]
\centering
\includegraphics[width=\textwidth]{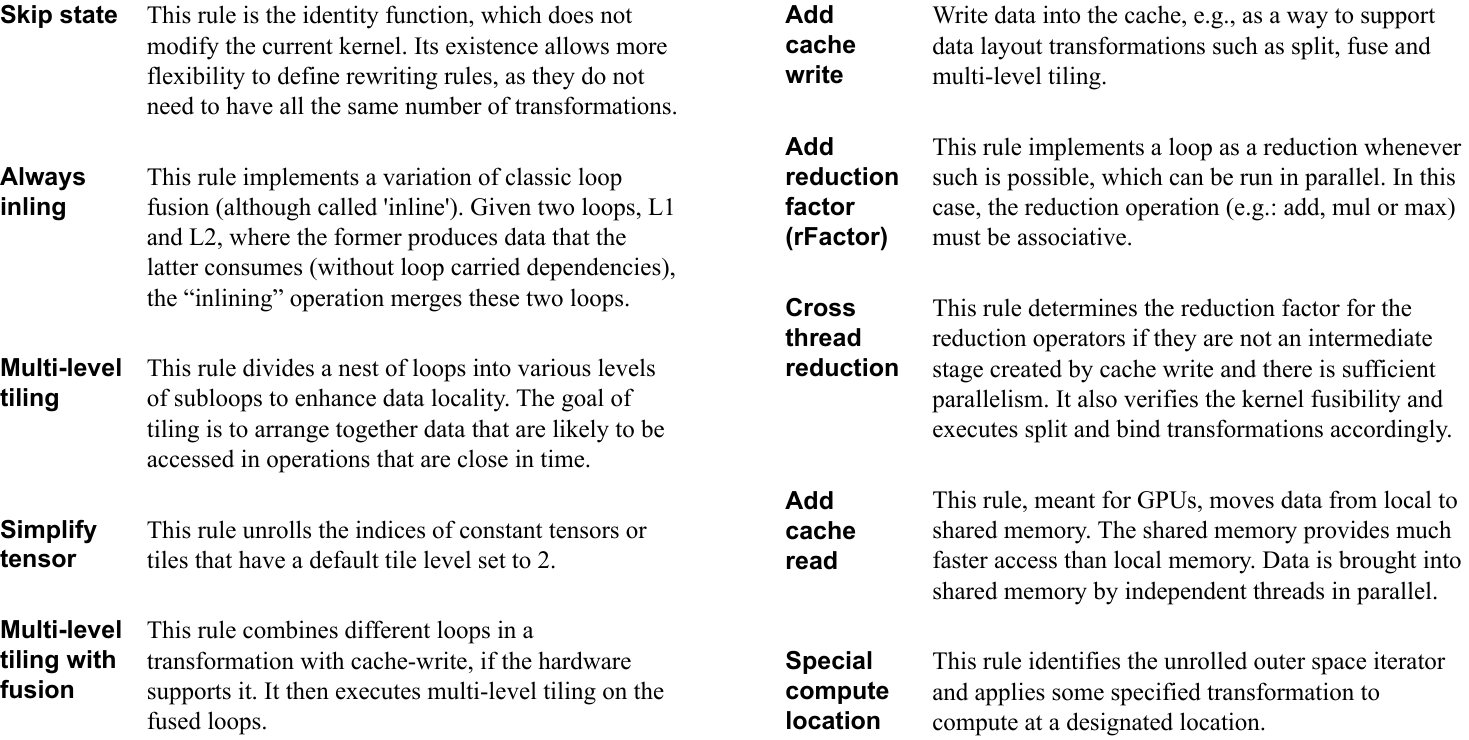}
\caption{Sketch generation rules.
\texttt{Ansor} uses these rules to change the kernel search space.
Each rule modifies a sketch, e.g., fusing, splitting or tiling loops.
However, these rules do not change the annotations in the sketch.}
\label{fig:rulesLayout}
\end{figure}

\paragraph{Sketch Generation}
The generation of new sketches happens via the application of a small collection of rewriting rules, which Figure~\ref{fig:rulesLayout}
enumerates\footnote{Rules implemented in \url{https://github.com/apache/tvm/tree/main/src/auto_scheduler/search_policy} on 02/01/2024}.
Each rewriting rule provokes a code transformation, such as tiling, parallelization, or unrolling.
The sketch generation rules are hardware-dependent: some rules, like those that bind loop iterations to threads, are only well-defined for GPUs, for instance.
Notice that in addition to the rules seen in Figure~\ref{fig:rulesLayout}, users can still add custom rules to the set of available sketch transformations.

\begin{figure}[hbt!]
\centering
\includegraphics[width=\textwidth]{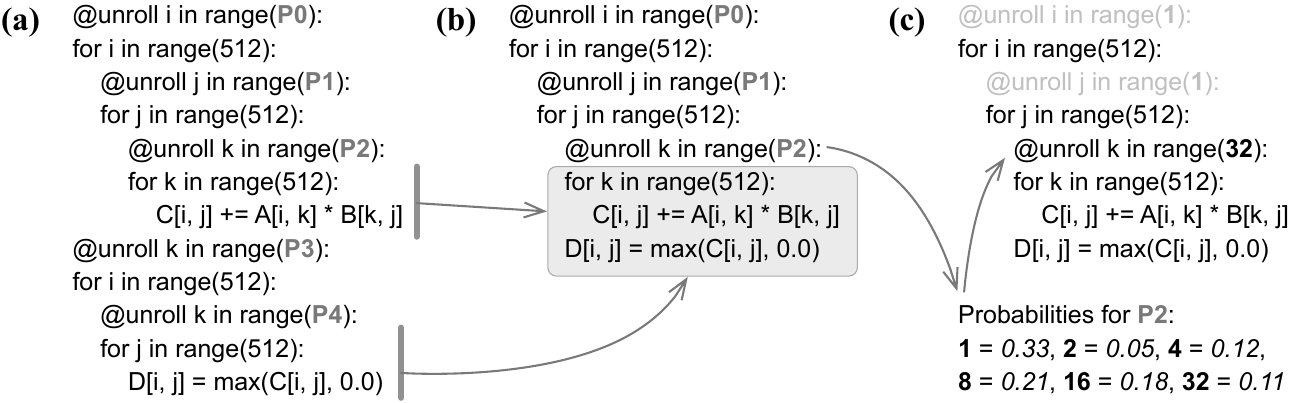}
\caption{(a) Sketch of the abstract kernel seen in Figure~\ref{fig:abstract_kernel} (a).
(b) Sketch that ensues from the application of the ``Always inlining'' rule.
(c) Sketch that ensues from the initialization of the unrolling factor.}
\label{fig:exSketchInit}
\end{figure}

\begin{example}
\label{ex:sketch}
Figure~\ref{fig:exSketchInit} (a) shows a variation of the sketch earlier seen in Figure~\ref{fig:abstract_kernel} (b).
Each loop has an unrolling factor, which will have to be annotated in the initialization phase of the autotuning process.
Figure~\ref{fig:exSketchInit} (b) shows an application of the ``Always inlining'' rule onto Figure~\ref{fig:exSketchInit} (a).
It is important to note that neither of these program representations qualifies as a ``kernel'', since they are not executable.
To transform them into concrete kernels, the annotation within the unrolling factor must be assigned an actual value.
\end{example}


\begin{figure}
\centering
\includegraphics[width=\textwidth]{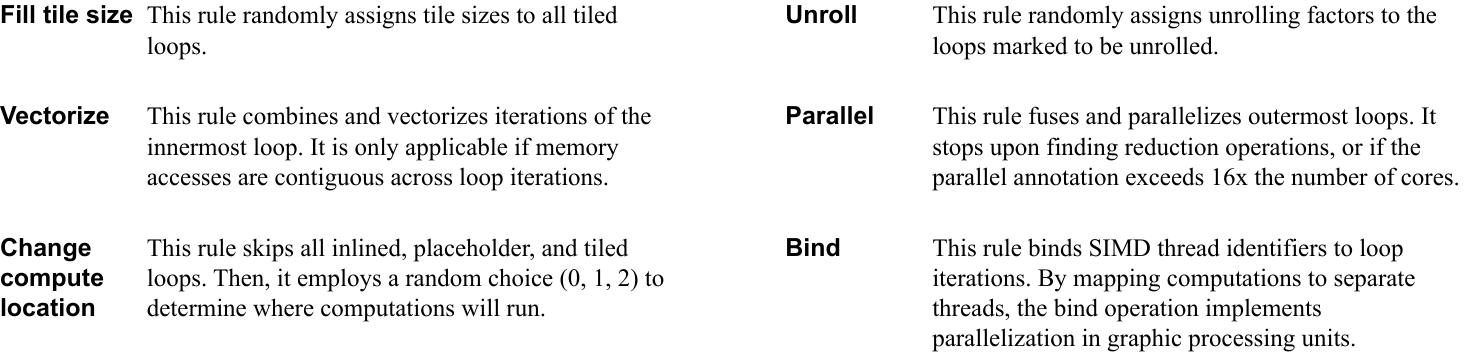}
\caption{Rules that \texttt{Ansor} uses to create an initial population of kernels, which will be the starting point to the evolutionary search.}
\label{fig:initializationRules}
\end{figure}

\paragraph{The Initialization of Annotations}
Sketches are not executable programs: they have annotations which must be replaced with actual values.
These values are the parameters of optimizations, as Definition~\ref{def:space} explains.
Thus, as a second step of the iterative exploration approach of \texttt{Ansor}, an initial population of annotated sketches is produced via the application of ``initialization rules''.
Figure~\ref{fig:initializationRules} summarizes the rules available in \texttt{Ansor} at the time this work was produced.
Each initialization rule is parameterized by a probability distribution, which associates concrete values with the probability that they can be chosen.
Example~\ref{ex:init} shows how these distributions are used.

\begin{example}
\label{ex:init}
Figure~\ref{fig:exSketchInit} (c) shows the concrete kernel that comes out of the application of the ``Unroll'' rule (in Figure~\ref{fig:initializationRules})
to the sketch in Figure~\ref{fig:exSketchInit} (b).
Figure~\ref{fig:exSketchInit} (c) also shows the probability distribution used to randomly choose the initial unrolling factor of 32.
\end{example}

\begin{figure}
\centering
\includegraphics[width=\textwidth]{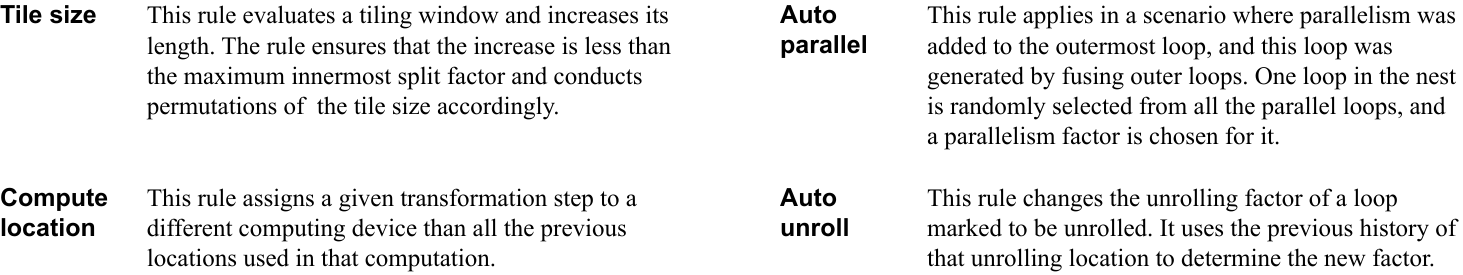}
\caption{Mutation rules for evolutionary search.
In contrast to the initialization rules seen in Figure~\ref{fig:initializationRules}, the mutation rules take the history of previous annotations when determining the next value of a given annotation.}
\label{fig:mutationRules}
\end{figure}

\paragraph{Evolution of the Annotated Sketch}
To explore different kernel spaces, \texttt{Ansor} keeps a population of promising {\it candidate kernels}.
This population is updated in an iterative process.
At each iteration, the current population of candidates evolves through the application of the mutation strategies enumerated in Figure~\ref{fig:mutationRules}.
To select the next set of candidates, \texttt{Ansor} does not run every kernel in the current population.
Instead, it uses a cost model to select candidates that are likely to run efficiently.
These promising candidates are executed, and the result of these samples is used to recalibrate the cost model; hence, improving the estimates of the next candidates.
Periodically, the algorithm reports statistical information regarding the search progress, including maximum and minimum scores, population size, and mutation success rates.
Upon completion of the specified number of iterations, the best-performing kernels are recorded as the output.

\paragraph{Termination in \texttt{Ansor}}
The number of possible schedules is very large; hence, \texttt{Ansor} limits the amount of schedules with a {\it budget of trials}.
Each trial consists of the observation of the execution of an actual schedule, which happens at the end of the evolutionary phase.
Because a machine learning model contains many kernels, an initial round of trials is partitioned among these layers.
Layers are grouped into a worklist, and receive a quota of trials in round-robin fashion.
After an initial round of optimizations, layers that run for a very short time are removed from this worklist.
This process ensures that layers that run for the longest time are subject to more extensive optimizations.
This approach is what \citet{Zheng20} call ``optimizing with gradient descent''.
Because the budget of trials is fixed, \texttt{Ansor} is guaranteed to terminate.

\paragraph{Limitations of \texttt{Ansor}}
The main limitation of \texttt{Ansor} is the fact that it is oblivious to the structure of the search space.
For instance, if we observe a performance improvement by increasing the unrolling factor of a loop from 4 to 6, it is likely that if we increase it further to 8, another improvement will also be observed.
However, if going to 8 results in performance degradation, then further increases are likely to not bring improvements either.
\texttt{Ansor}'s exploitation approach, via an evolutionary algorithm, is not aware of this notion of neighborhood between kernels or of potential convex regions in the optimization space.

\subsection{Space Exploitation via Droplet Search}
\label{sub:droplet_exp}

\begin{figure}[ht]
\centering
\includegraphics[width=\textwidth]{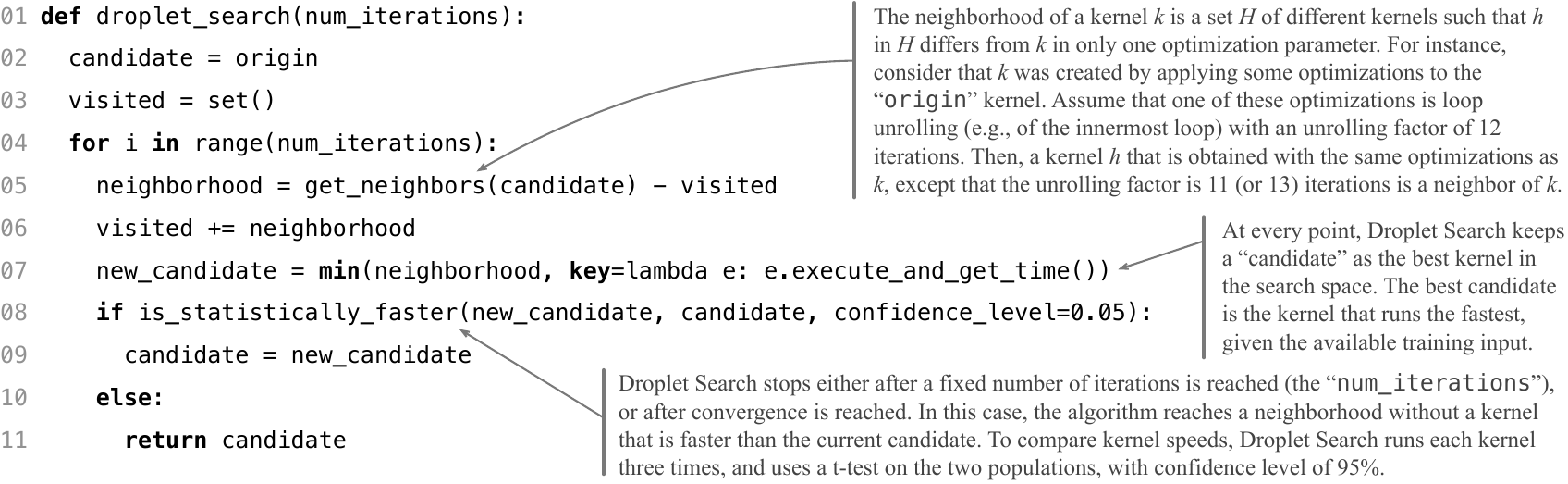}
\caption{The Droplet Search kernel scheduling algorithm.
This pseudo-code is a simplified version of the original presentation of the algorithm, taken from~\cite{Canesche24}.
We have removed speculation and parallelism from this version as
these features are immaterial for the presentation of our ideas.}
\label{fig:simplifiedDropletSearch}
\end{figure}

Droplet Search~\cite{Canesche24} is a kernel scheduling algorithm available in \texttt{AutoTVM}.
\texttt{AutoTVM} differs from \texttt{Ansor} because it does not create new sketches.
Rather, it is restricted to modifying the parameters of a single sketch---the {\it origin} of the optimization space.
\texttt{AutoTVM} provides several independent scheduling approaches: random sampling, grid sampling, genetic sampling, etc.
However, only Droplet Search will be of interest to this presentation\footnote{Section~\ref{sub:search_technique} shall compare Droplet Search with the other techniques available in \texttt{AutoTVM} as when they are used as \texttt{Ansor}'s exploitation mechanism.}.
Droplet Search is a variation of an exploitation algorithm called {\it Coordinate Descent}\footnote{It is unclear who invented Coordinate Descent. Descriptions of the algorithm can be found in classic textbooks~\cite{Zangwill69}. For a comprehensive overview, we recommend the
work of \citet{Wright15}.}.
It relies on the premise that the parameters of a sketch can be arranged into a coordinate space.
Figure~\ref{fig:simplifiedDropletSearch} contains an annotated version of the algorithm.

\begin{example}
\label{ex:droplet}
Let us assume a sketch formed by two optimizations: unrolling and tiling.
For the sake of this example, unrolling supports five ``unrolling factors'': $\{1, 2, 3, 4, 5\}$.
These are the parameters of the loop unrolling optimization.
Tiling is parameterized by the size of the tiling window.
Let us assume the following sizes: $\{1, 2, 4, 8, 16\}$.
The optimization space, in this case, is formed by $5 \times 5$ points, such as $(1, 1)$, which means no optimization, or $(3, 16)$, which indicates that the loop must be unrolled three times, and then tiled with a window of size 16.
These points, e.g., $(1, 1)$, $(3, 16)$, etc, are the {\it coordinates} of the optimization space.
\end{example}

From the notion of coordinates, Droplet Search defines a {\it neighborhood function}: a function that returns the {\it neighbors} of a given coordinate.
Intuitively, the neighbors of a coordinate are the points that are the closest to it.
In Example~\ref{ex:droplet}, the neighbors of $(\mbox{unrolling}=3, \mbox{tiling}=8)$ would be the points $(2, 8)$, $(4, 8)$, $(3, 4)$ and $(3, 16)$.
From this concept of neighborhood, Droplet Search works iteratively as follows:

\begin{enumerate}
\item At iteration zero, let the best current candidate be the set of parameters that implement no optimization.
\item Let $(c_1, c_2, \ldots, c_n)$ be the best set of parameters discovered up to iteration $i$.
\begin{enumerate}
\item If there exists $c_i', 1 \leq i \leq n$, such that $(c_1, \ldots, c_i', \ldots, c_n)$ yields a faster kernel than $(c_1, \ldots, c_i, \ldots, c_n)$, then update the current best candidate to use $c_i'$ instead of $c_i$.
\item If there is no such $c_i'$, then the search terminates.
\end{enumerate}
\end{enumerate}

\paragraph{Limitations of Droplet Search}
Droplet search is a fast search algorithm when compared to \texttt{Ansor} or to other approaches available in \texttt{AutoTVM}~\cite{Canesche24}.
However, it has two fundamental limitations:

\begin{itemize}
\item Droplet Search is restricted to a single sketch.
In other words, it can modify a sketch's annotations, but it cannot create new sketches.
This is a limitation of any search algorithm used in \texttt{AutoTVM}, but it is not a limitation of \texttt{Ansor}.

\item Droplet Search depends highly on the initial schedule it receives as the {\it seed} of the search procedure.
If this initial schedule does not exist in the same convex region as the optimal schedule, then Droplet Search cannot find the optimal schedule.
\end{itemize}
By combining \texttt{Ansor} and Droplet Search, we hope to circumvent the limitations of both search techniques.
Section~\ref{sub:combination} explains how these two approaches can be used together.

\subsection{Combining \texttt{Ansor} with Droplet Search}
\label{sub:combination}

\begin{figure}[th]
\centering
\includegraphics[width=\textwidth]{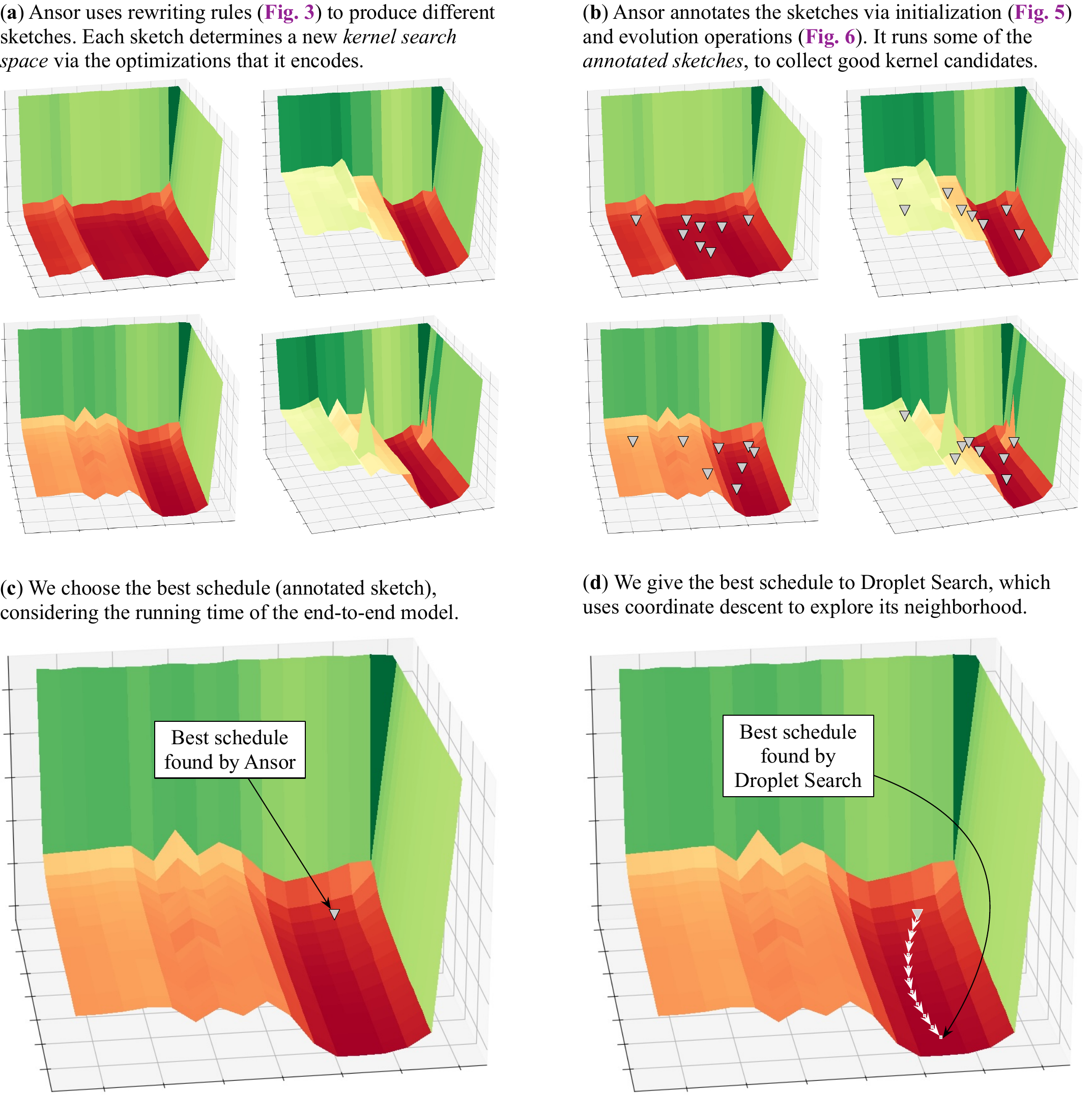}
\caption{Coarse exploration of different kernel search spaces with \texttt{Ansor}, and careful exploitation of best candidate with Droplet Search.}
\label{fig:combined_intuition}
\end{figure}

To combine Droplet Search and \texttt{Ansor}, we determine two parameters:
\begin{itemize}
\item $K$: the budget of trials of \texttt{Ansor}.
\item $N < K$: a subset of trials.
\end{itemize}
We then proceed as follows:
\begin{enumerate}
\item Run \texttt{Ansor} on the target model using only $N$ trials.
\item Give the best schedule found with $N$ trials to Droplet Search.
\item Run Droplet Search up to convergence.
\end{enumerate}
Figure~\ref{fig:combined_intuition} provides some intuition on this {\it modus operandi}.
As the figure illustrates, the proposed technique seeks to use \texttt{Ansor} to explore the universe of sketches and then use Droplet Search as the core strategy to explore concrete representations of these sketches.

In Section~\ref{sec:results}, we demonstrate that by choosing proper values for $K$ and $N$, we can outperform \texttt{Ansor} in two ways: first, producing faster end-to-end machine learning models; second, reducing the search time of \texttt{Ansor}.
The modifications needed to add this combination to the current code base of Apache TVM are relatively small: up to 307 lines of code in the \texttt{TVM} repository (Release 17, Apache TVM v0.15.0).

\section{Evaluation}
\label{sec:results}

This section evaluates the idea proposed in this paper. In particular, it seeks to demonstrate that by exploiting, via Droplet Search, a reduced set of samples explored by \texttt{Ansor}, it is possible to outperform \texttt{Ansor} itself. We shall refer to this new version of \texttt{Ansor}, which uses Droplet Search, as the \textit{Combined Approach}. Henceforth, we denote it as \texttt{DPAnsor}, reserving \texttt{Ansor} for the original implementation of that tool. In what follows, we explore six research questions:

\begin{description}
\item[RQ1:] How many samples does \texttt{DPAnsor} need to observe to produce kernels that outperform those produced by \texttt{Ansor} with 10,000 trials?
\item[RQ2:] How many samples can \texttt{DPAnsor} observe and still outperform \texttt{Ansor} in terms of search time when the latter uses 10,000 trials?
\item[RQ3:] How does the size of models impact the behavior of \texttt{DPAnsor}, in terms of kernel performance and search speed?
\item[RQ4:] How does Droplet Search compare to other search techniques available in \texttt{AutoTVM}, in terms of their ability to exploit \texttt{Ansor}'s results?
\item[RQ5:] How does the average number of samples that Droplet Search gauges per layer vary with the initial budget allocated to \texttt{DPAnsor}'s exploration phase?
\item[RQ6:] How does \texttt{DPAnsor} compare with \texttt{PyTorch} and
\texttt{Ansor} in terms of the speed of the kernels that it produces,
considering a range of different machine-learning kernels?
\end{description}

Before diving into the research questions, we explain our experimental setup.
Notice that a fully containerized version of this methodology has been organized as a \texttt{docker} image, which is publicly available at \url{https://github.com/lac-dcc/bennu}.

\begin{figure}[ht]
    \centering
    \includegraphics[width=1.0\textwidth]{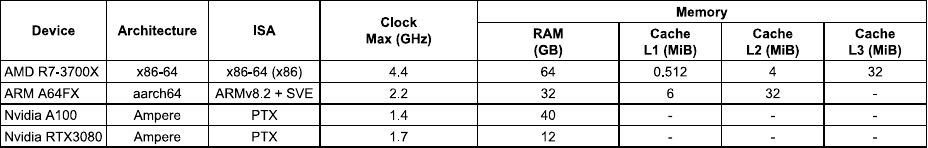}
    \caption{The architectures evaluated in this report.}
    \label{fig:device}
\end{figure}

\paragraph{Hardware and Software.}
We evaluated the scheduling approaches on four different architectures, as shown in Figure~\ref{fig:device}.
The hardware consists of a general-purpose desktop architecture (AMD Ryzen 7~\cite{amdRyzenR7}), a cluster-based machine (ARM A64FX~\cite{ookami}), and two graphics processing units (NVIDIA A100~\cite{nvidiaA100} and NVIDIA RTX3080~\cite{rtx3080}).
The experiments reported in this section use versions of \texttt{Ansor} and \texttt{AutoTVM} (Droplet Search) available at \texttt{Apache TVM v0.13.0}, released in July 2023. The version used for \texttt{TensorFlow} was 2.14, and \texttt{PyTorch} was 2.0+cu118.

\paragraph{Benchmarks.}
This section evaluates kernel scheduling across twenty neural networks.
The first column of Figure~\ref{fig:res_x86} contains the complete list of these models.
All these models are implemented using the \texttt{ONNX} representation to make a comparison between \texttt{TVM},
\texttt{PyTorch} and \texttt{TensorFlow} possible, as seen in Section~\ref{sub:micro_kernels}.
The models used in our study are sourced from the ONNX model zoo available at \url{https://github.com/onnx/models}.

\paragraph{Methodology.}
A machine learning model forms a graph of {\it kernels}.
\texttt{Ansor} optimizes machine learning models per kernel, assuming kernels can be independently optimized.
It starts with a {\it budget of trials}, where each trial is a transformation that can be applied to a kernel.
Let us call this budget $K$.
\texttt{Ansor} ensures that each kernel receives a fraction of these $K$ trials.
Currently, this initial fraction is $\mathit{min}(K/L, 64)$, where $L$ is the number of layers (kernels) in the model.
After an initial round of optimizations, \texttt{Ansor} applies the remaining trials onto kernels that run for the longest time.
This approach directs the optimization effort to the kernels that are more likely to contribute to the overall running time of the end-to-end model.
In what follows, all the results we report are relative to a baseline version of \texttt{Ansor} equipped with a budget of 10,000 trials.
We shall test \texttt{DPAnsor} with either $K = 1, 10, 25, 50, 100, 200, 300$, or $1,000$ trials.
Suppose we choose $K = 100$, for instance. In that case, we will run \texttt{Ansor} with a budget of 100 trials, pick the best configuration (which results from the independent optimization of the kernels), and give this configuration to \texttt{AutoTVM}'s Droplet Search.
We then let Droplet Search run until it reaches convergence.

\paragraph{Confidence}
Sections~\ref{sub:kernel_quality}-\ref{sub:search_technique} show relative results.
It could be possible that the running times presented for kernels and schedulers is similar enough to the point of hindering our conclusions meaningless.
However, such is not the case.
Scheduling runs for a very long time.
To give the reader an idea, \texttt{Ansor}, with a budget of 10,000 trials (or baseline), takes 11,195 seconds to schedule \texttt{AlexNet}, our smallest model.
This time drops to 2,839 seconds considering \texttt{DPAnsor} with a budget of 300 trials.
The running time of kernels is faster: end-to-end models run for a few seconds.
However, in every experiment, we consider averages of three samples and only report speedups if the p-value (produced via the non-parametric Wilcoxon rank-sum test) for the difference between the two populations is below 0.01.

\subsection{RQ1 -- On the Quality of End-to-End Models}
\label{sub:kernel_quality}

We consider that a version of an end-to-end model is better than another for a given architecture when it runs faster in that architecture.
The execution time of a model is determined by the schedule of the kernels that constitute it.
If we apply Droplet Search to the best model produced by \texttt{Ansor} after it observes 10,000 trials, we will likely improve the model (at least, we should not make it worse).
However, this section shows that obtaining a better model via \texttt{DPAnsor} with a much lower budget is possible in four different architectures.
For brevity, we analyze in details results obtained on an x86 CPU.
For the other architectures, we show only summarization results.

\begin{figure}[ht!]
    \centering
    \includegraphics[width=\textwidth]{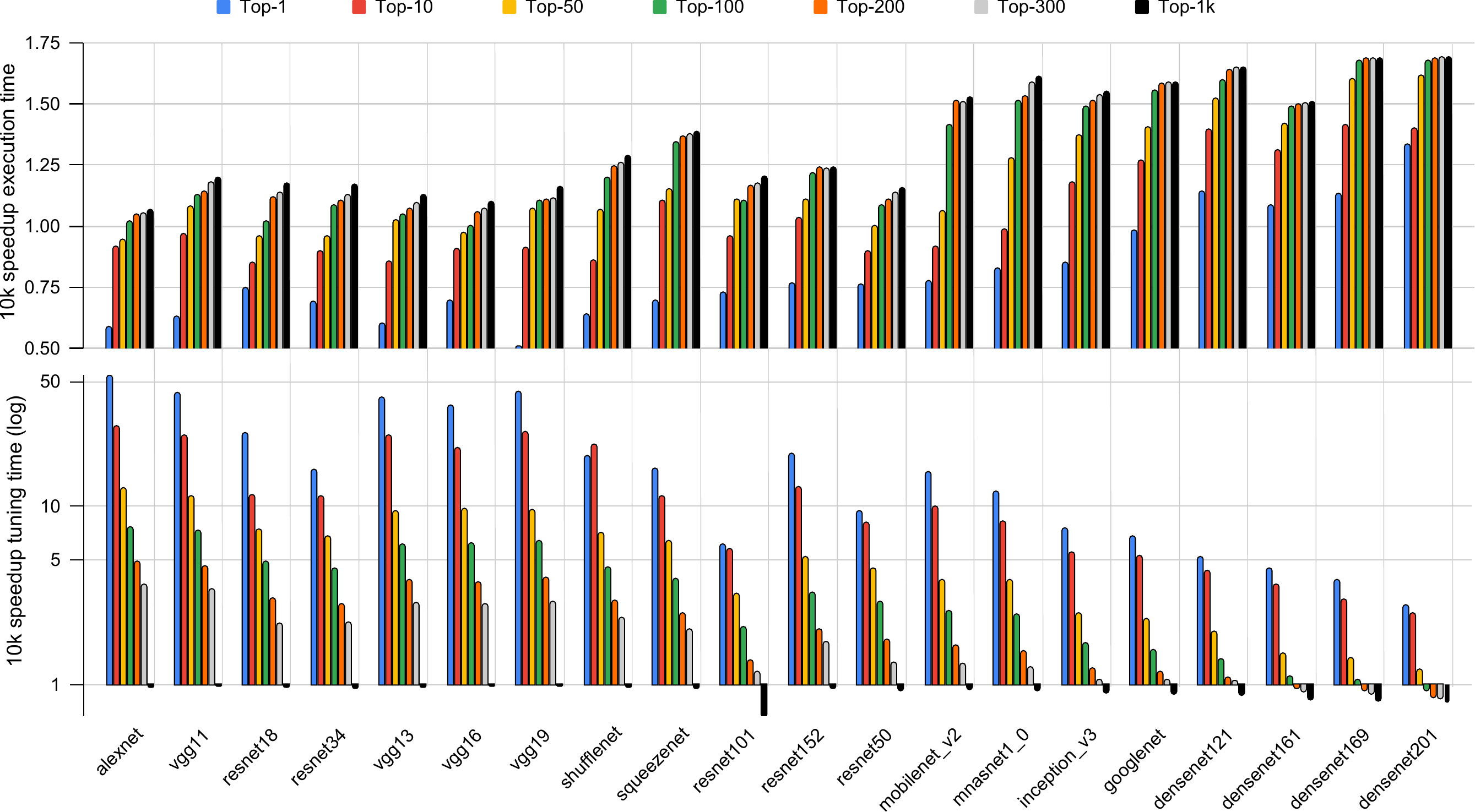}
    \caption{Comparative Analysis of Optimization Results on the x86 Architecture Using an AMD Ryzen 7 3700X Processor.
    Numbers show $\mathtt{Ansor}/\mathtt{DPAnsor}$ ratios.
    Thus, results higher than 1.0 (in blue) denote improvements of \texttt{DPAnsor} (this paper) over \texttt{Ansor}.}
    \label{fig:res_x86}
\end{figure}

\paragraph{Discussion: AMD Ryzen 7 (x86-64)}
The x86 architecture represents a widely adopted instruction set architecture (ISA), serving as the basis for Intel and AMD processors.
Figure \ref{fig:res_x86} compares kernel speed and search time of models running on an x86-64 CPU.
The top part of the figure (labeled ``10k speedup execution time'') compares the running time of the kernels produced by \texttt{Ansor} and \texttt{DPAnsor}.
The lower part (labeled ``10k speedup tuning time'') compares the search time of these two approaches, and shall be discussed in Section~\ref{sub:search_time}.
In both cases, bars above 1.0 denote improvements of \texttt{DPAnsor} over
\texttt{Ansor}.
In terms of kernel speed, sampling 25 configurations with \texttt{DPAnsor}
is sufficient---in most models---to outperform \texttt{Ansor} with 10K samples.
Speedups improve gradually as more samples are added to \texttt{DPAnsor}, to the point that with 1,000 samples, we see an average speedup (geometric mean) of 34\%.


\begin{figure}[ht]
    \centering
    \includegraphics[width=\textwidth]{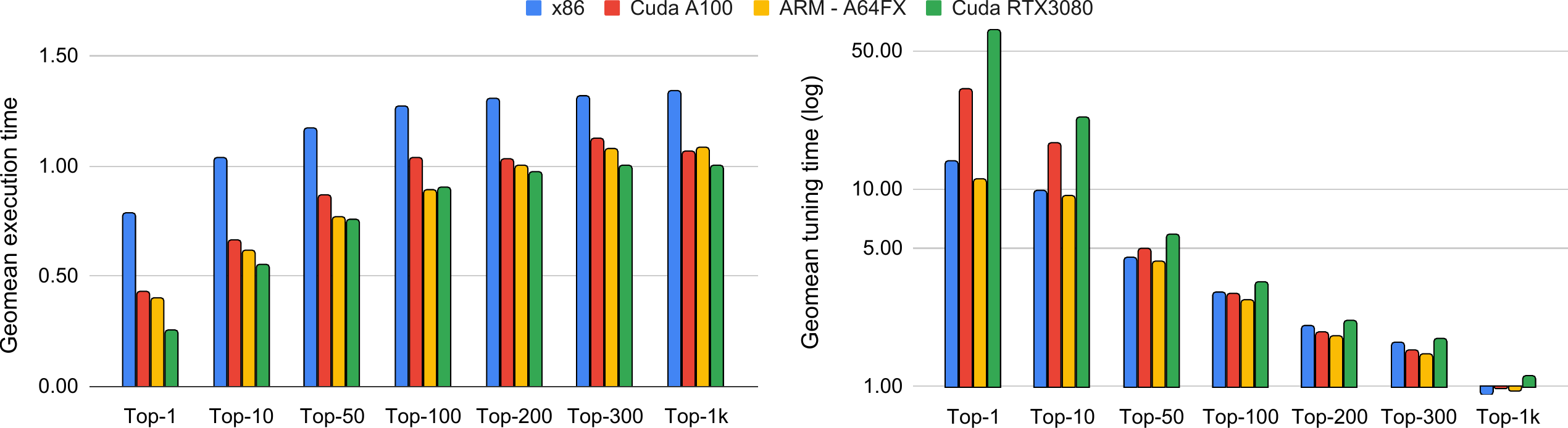}
    \caption{Left: relative speed of kernels produced by \texttt{DPAnsor} over
    \texttt{Ansor} on four different architectures.
    Right: relative search time of \texttt{DPAnsor} over \texttt{Ansor}.
    Every bar is the geometric mean of relative times observed on 20 different
    models.
    Results above 1.0 represent improvements of \texttt{DPAnsor} over
    \texttt{Ansor}.}
    \label{fig:results_geomean}
\end{figure}

\paragraph{Discussion: Other Architectures}
Figure~\ref{fig:results_geomean} (Left) compares the relative speed of kernels produced by \texttt{DPAnsor} over \texttt{Ansor} in all the four architectures available for this study: in addition to the AMD x86 of Figure~\ref{fig:res_x86}, we see an Nvidia A100(Ampere), an Nvidia RTX3080 (Ampere), and an ARM A64FX (aarch64).
In every case, speedups of \texttt{DPAnsor} relative to \texttt{Ansor} emerge
consistently with $K=300$.
However, with $K=100$, we have already recorded speedups on the Nvidia A100 and on the ARM A64.
We failed to see meaningful speedups on the Nvidia RTX3080 because \texttt{Ansor}, with a budget of 10K samples, seems to be very close to achieving peek performance on this GPU.
Even if we increase its budget of samples, we could not obtain better kernels
to the 20 different models used in this study.
Nevertheless, notice that \texttt{DPAnsor} achieves this peek performance with 300 samples (plus a few---less than 50---samples probed by Droplet Search).
On the A100 GPU, in contrast, after 300 samples, \texttt{DPAnsor} already delivers kernels 12\% faster than those produced by \texttt{Ansor}.




\subsection{RQ2 -- On the Search Time}
\label{sub:search_time}

We define a scheduling approach as faster than another if it requires less time to converge to an end-to-end model's final, optimized version. The search time of \texttt{Ansor} encompasses the time spent applying optimizations to kernels, deriving new optimizations, and running the kernels themselves, with a limit set at 10,000 trials. On the other hand, the search time of \texttt{DPAnsor} involves all the steps of \texttt{Ansor}, constrained to a lower number of trials, along with the time it takes to run Droplet Search until convergence on the kernels that compose a model. Although we have restricted Droplet Search to a maximum of 100 trials, it typically converges well before reaching that limit. This section compares the search time between \texttt{Ansor} and \texttt{DPAnsor}.

\paragraph{Discussion}
The lower part of Figure~\ref{fig:res_x86} compares search times on x86.
For most models, \texttt{DPAnsor} is consistently faster than \texttt{Ansor} for any number of trials up to $K = 300$.
At 1,000 trials, \texttt{Ansor} becomes consistently faster.
Also, \texttt{Ansor} tends to outperform \texttt{DPAnsor} for very large models.
This fact happens due to the longer time that Droplet Search takes to converge: the more complex the model, the more room Droplet Search will have to optimize it.
Section~\ref{sub:model_size} further discusses the impact of the model size on the behavior of \texttt{DPAnsor}.
Figure~\ref{fig:results_geomean} summarizes the search time comparison for the other architectures.
The pattern is similar to the one observed on x86: \texttt{DPAnsor} is consistently faster when $K \leq 300$, and slower (except on the Nvidia RTX3080) at $K=1,000$.

\subsection{RQ3 -- On the Impact of Model Size}
\label{sub:model_size}

The behavior of \texttt{DPAnsor}, when compared to \texttt{Ansor}, varies with the model's size.
We summarize this variation with two observations:
\begin{enumerate}
\item The larger the model, the less samples \texttt{DPAnsor} needs to observe to outperform \texttt{Ansor}, if \texttt{Ansor} uses a budget of 10,000 samples.
\item The larger the model, the lower the benefit, in terms of search time, of \texttt{DPAnsor} over \texttt{Ansor}.
\end{enumerate}
The rest of this section provides data to support these two conclusions.

\begin{figure}[ht]
    \centering
    \includegraphics[width=\textwidth]{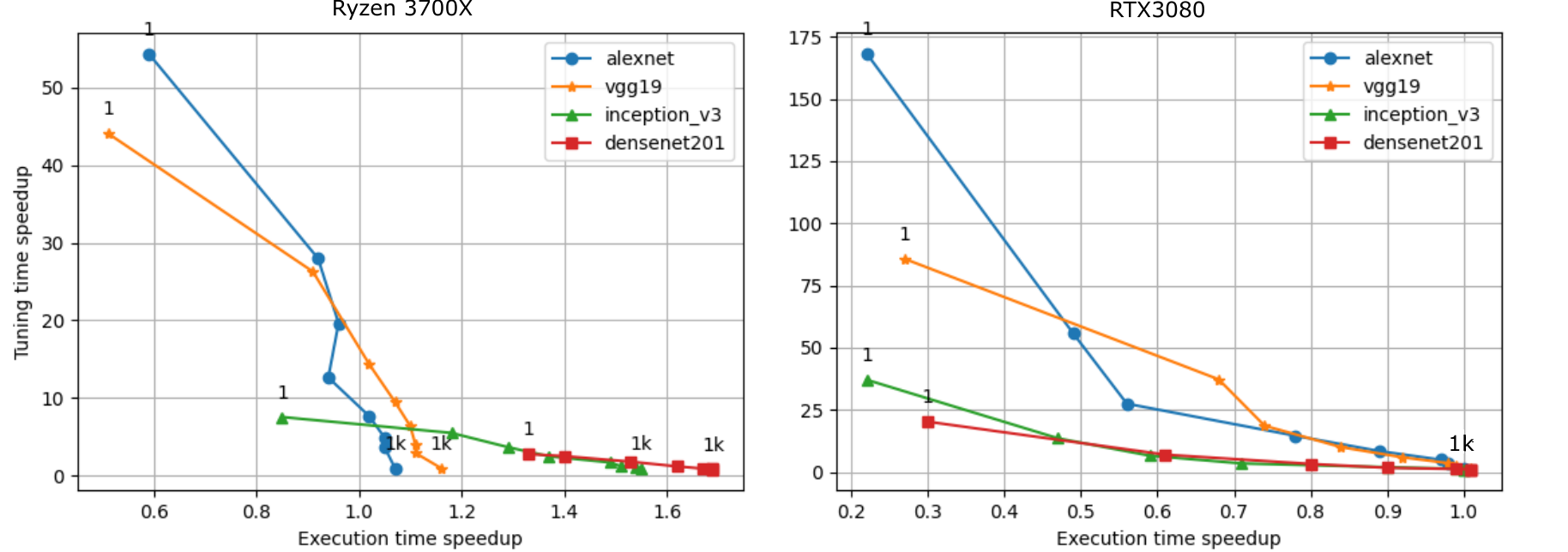}
    \caption{The search vs quality line that characterizes four models optimized in the AMD (Left) and in the NVIDIA (Right) setting: the larger the model, the lower the slope.
    Numbers on the X and Y axes are speedup/slowdown relative to \texttt{Ansor} with a budget of 10,000 trials.
    Numbers for the left chart are available in Figure~\ref{fig:res_x86}, and numbers for the right chart were observed on the Nvidia RTX3080.
    The labels on the dots (1 and 1k) refer to the number of trials given to \texttt{DPAnsor}.}
    \label{fig:behavior}
\end{figure}

\paragraph{Discussion: The Search vs Quality Slope}
The kernel optimization technique impacts two core numbers: the search time and the speed of the final model.
We can use these two quantities---search time (S) and model performance (P)---to define an $S \times P$ line characterizing the behavior of the optimization technique.
Figure~\ref{fig:behavior} shows these lines regarding four models and two architectures: AMD's x86 and NVIDIA's Ampere.
We chose these two architectures because x86 is the scenario where \texttt{DPAnsor} performs better, and Ampere is the scenario where it performs worse.

Figure~\ref{fig:behavior} uses our two smallest and two largest models.
The numbers on the axes show ratios between \texttt{DPAnsor} and \texttt{Ansor}; the latter using a budget of 10,000 trials.
Each dot in Figure~\ref{fig:behavior} refers to the number of trials that \texttt{DPAnsor} is allowed to observe before shifting to Droplet Search.
The figure labels dots that refer to \texttt{DPAnsor} with one sample (its most restrictive scenario) and dots that refer to 1,000 samples (its least restrictive scenario).

The slopes of the lines in Figure~\ref{fig:behavior} are always negative, meaning that as more samples are given to \texttt{DPAnsor}, the difference between its search time and \texttt{Ansor}'s reduces, but the quality of the kernels that it finds improves.
However, the inclination changes with the size of the model.
The larger the model, the lower the benefit of \texttt{DPAnsor} over \texttt{Ansor} in terms of search time; but the higher the relative benefit in terms of kernel speed.
This result is due to \texttt{Ansor}'s fixed budget of 10,000 trials.
In a small model, more trials are distributed to each layer; in a large model, each layer receives only a handful of trials.

\begin{figure}[ht]
    \centering
    \includegraphics[width=1\textwidth]{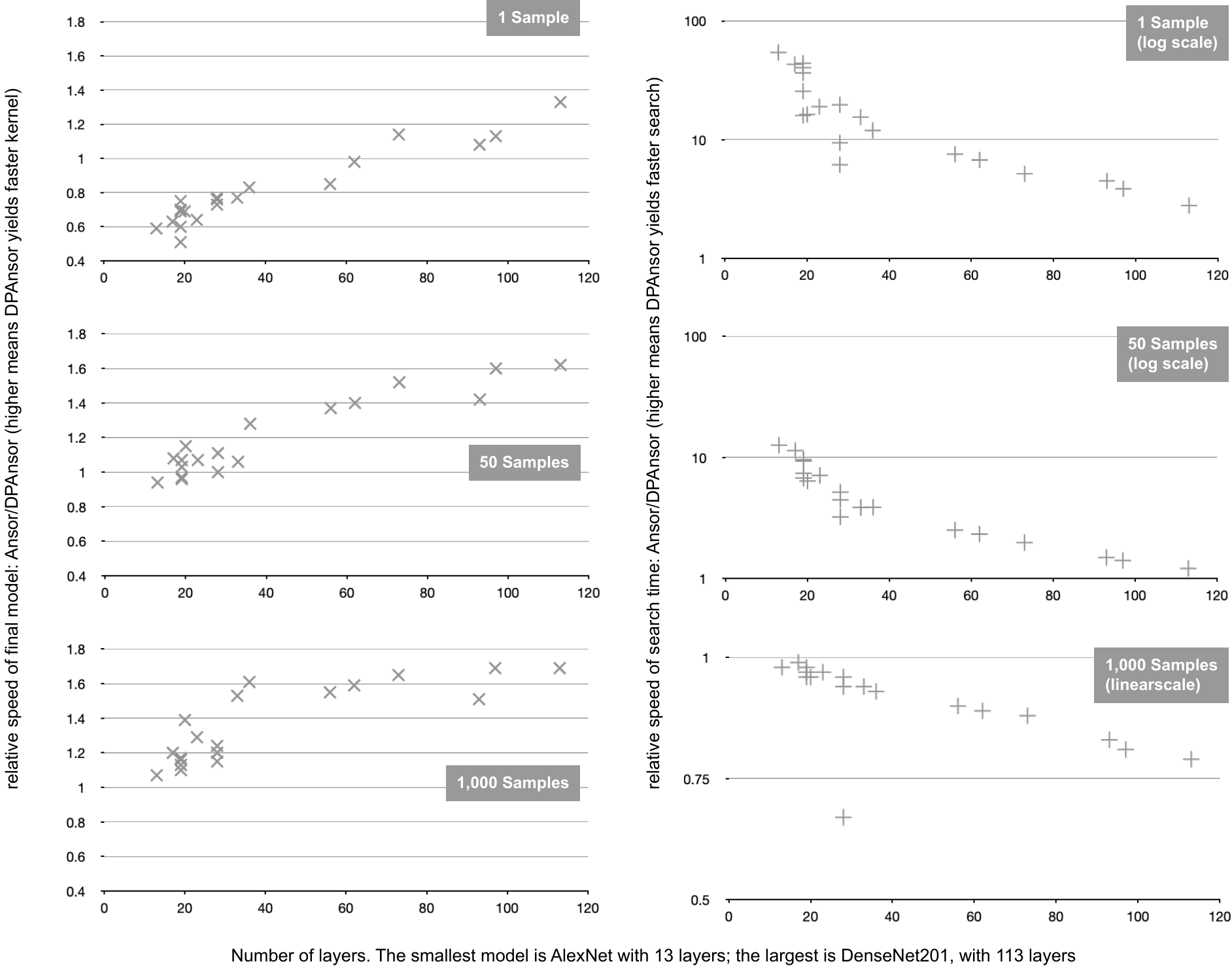}
    \caption{(Left) Variation in kernel quality ($\mathtt{Ansor}/\mathtt{DPAnsor}$)) with model size.
    (Right) Variation in search time ($\mathtt{Ansor}/\mathtt{DPAnsor}$)) with model size.}
    \label{fig:modelSizeImpact}
\end{figure}

Figure~\ref{fig:modelSizeImpact} provides further data that supports the previous observations.
The figure shows how kernel quality and search time vary with the size of models.
If we consider \texttt{AlexNet}, which has only 13 layers, each layer might receive, on average, $10^4/13$ trials.
Inevitably, good kernels will emerge from this search.
Thus, the benefit of the exploitation phase, which uses Droplet Search, tends to be smaller.
On the other hand, if we consider \texttt{DenseNet201}, which has 113 layers, then \texttt{Ansor} allocates, on average, $10^4/113$ trials per layer---less than 100 samples per each kernel of the computational graph.
This number is too low to effectively explore the space of possible kernel implementations.
In this case, Droplet Search has more opportunity to improve the kernels that \texttt{Ansor} finds.
Even using only one trial to find the origin of the optimization space is already enough to have \texttt{DPAnsor} outperforming \texttt{Ansor} in terms of the quality of the model.

\subsection{RQ4 -- On the Search Technique}
\label{sub:search_technique}

The core idea of this paper consists in using Droplet Search to exploit results produced by \texttt{Ansor}.
Therefore, an immediate question from this proposal is: {\it what if other search techniques are used instead of Droplet Search, as the basic exploitation technique}?
Indeed, \texttt{AutoTVM}, the framework hosting Droplet Search, provides four other search techniques that could fill the same role as Droplet Search.
These techniques are described as follows:
\begin{description}
\item [Random:] a random search on the space of valid optimization parameters.
In this case, each sample is randomly produced.
Search is stateless, meaning that the results of a kernel bear no influence on the choice of the next kernel.
\item [Grid:] a grid search on the space of valid parameters of the optimizations.
In this case, each sample derives from a regular and exhaustive variation of each optimization parameter.
Search keeps a minimum of state, namely, a counter per optimization parameter, that goes over the range of acceptable values.
\item [GA:] a genetic algorithm that treats the sequence of optimization parameters as chromosomes.
Search is stateful: the running times of already seen kernels provide information to guide the synthesis of the next kernels via operations such as mutation, crossover and pruning.
\item [XGB:] search is based on gradient boosting, as implemented by the \texttt{XGBoost} Library~\cite{Chen16}.
Similar to the genetic algorithm, the search is stateful, as the running time of kernels guides the construction of the search tree.
\end{description}
In the rest of this section, we analyze the behavior of \texttt{DPAnsor}, once its search technique (the ``\texttt{DP}'' part of \texttt{Ansor}) is replaced by each one of the other four algorithms available in \texttt{AutoTVM}.

\paragraph{Methodology}
Every experiment in this Section allocates a budget of 300 samples to \texttt{Ansor} for each model that we autotune.
Exploitation, via one of the different search techniques in \texttt{AutoTVM}, uses a budget of 100 trials to optimize each layer of \texttt{AlexNet}.
Droplet Search converged before 100 trials in every layer, but all the other approaches run 100 samples, as they do not have a notion of convergence.
Thus, each other approach samples 1,200 kernel configurations.
Each search technique in \texttt{AutoTVM} exploits the best sequence of optimizations found by \texttt{Ansor}, which is the same for all of them.
Following the methodology adopted in Section~\ref{sub:model_size}, we report results for the AMD x86 and the NVIDIA 3080 boards.
These are the best and worst scenarios observed for \texttt{DPAnsor} in Sections~\ref{sub:kernel_quality} and~\ref{sub:search_time}.

\begin{figure}[ht]
    \centering
    \includegraphics[width=1\textwidth]{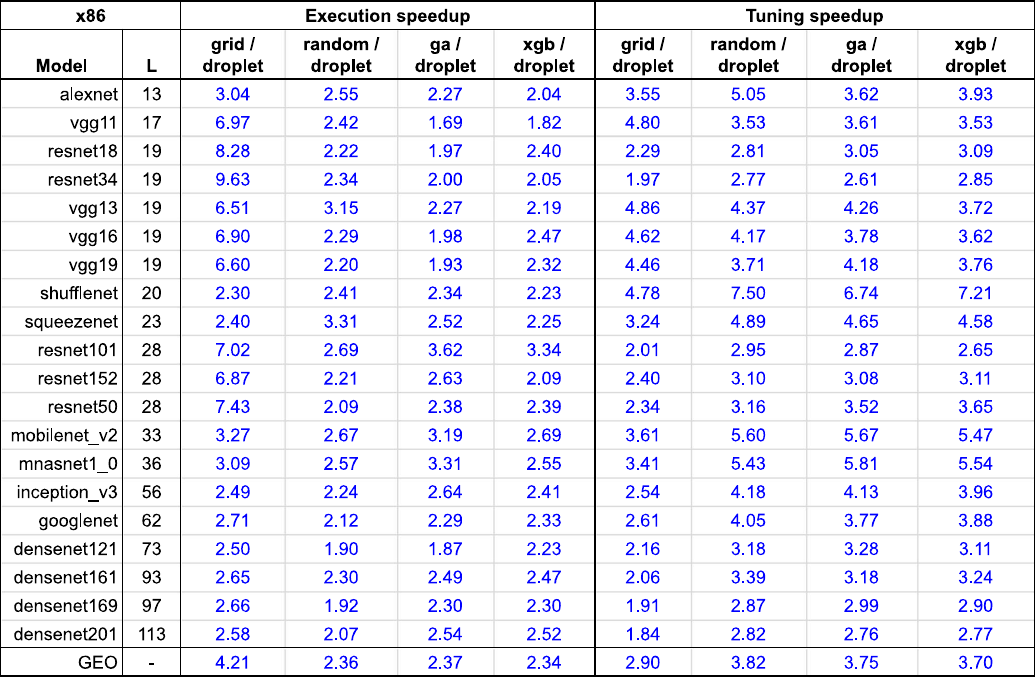}
    \caption{Comparison of different exploitation techniques used in tandem with \texttt{Ansor} on the x86-64 setting.
    Numbers are ratio of ``$\mathit{Search Technique}/\mathit{Droplet Search}$''.
    Thus, numbers above 1.0 demonstrate the effectiveness of the technique proposed in this paper.}
    \label{fig:comparison_droplet_x86}
\end{figure}

\paragraph{Discussion: AMD Ryzen 7 (x86-64)}
Figure~\ref{fig:comparison_droplet_x86} compares Droplet Search against other \texttt{AutoTVM} methods on the x86 architecture.
As an exploitation technique, Droplet Search demonstrates superior performance in terms of execution time and kernel speedup.
The \textbf{random} search technique is completely oblivious to the seed best provided by \texttt{Ansor}.
The \textbf{grid} search is almost oblivious to it: although the grid search starts from a likely good kernel, it diverges quickly from that configuration.
The other techniques can benefit from \texttt{Ansor}'s exploration phase.
We feed both \textbf{GA} and \textbf{XGB} with a promising seed; however, these search techniques still need random seeds to diversify the initial population of kernels from where they depart.
Because the other search techniques lack a convergence criterion, the relative performance of Droplet Search---in turning time---is even better: it is approximately 3x faster.

\begin{figure}[ht]
    \centering
    \includegraphics[width=1\textwidth]{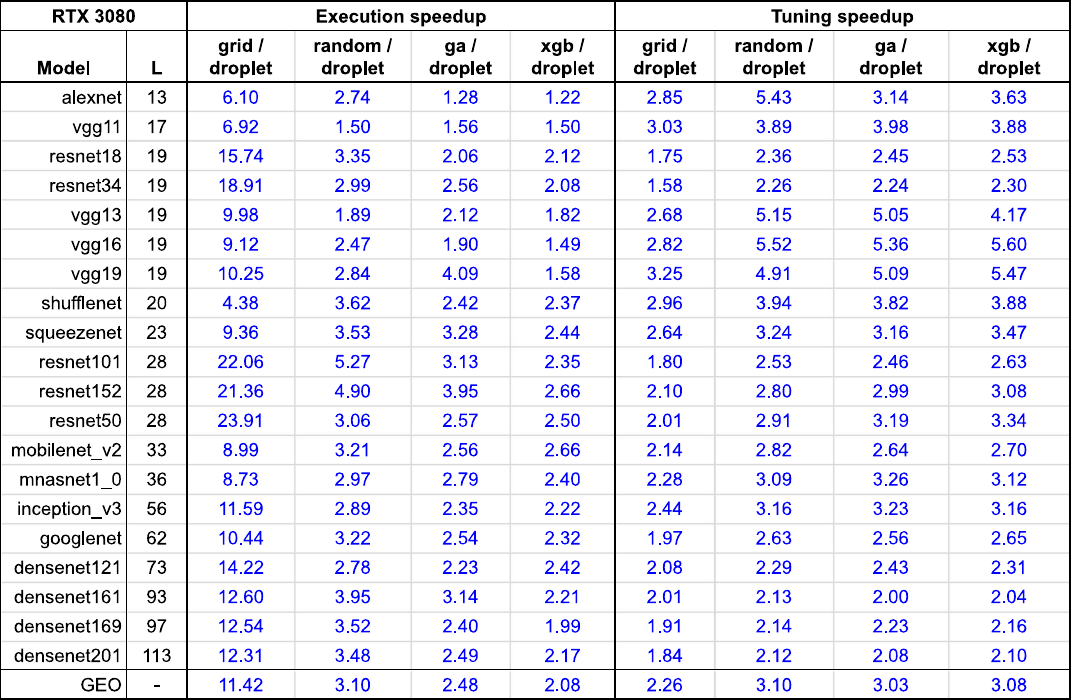}
    \caption{Comparison of different exploitation techniques used in tandem with \texttt{Ansor} on the Cuda setting.
    Numbers are ratio of ``$\mathit{Search Technique}/\mathit{Droplet Search}$''.
    Thus, numbers above 1.0 demonstrate the effectiveness of the technique proposed in this paper.}
    \label{fig:comp_cuda}
\end{figure}

\paragraph{Discussion: CUDA RTX 3080 (Ampere)}
In the GPU setting, Droplet Search is still a superior exploitation technique in terms of search time and kernel speed, as
Figure~\ref{fig:comp_cuda}  demonstrates.
In contrast to the CPU setting, invalid kernels are common in the GPU scenario.
\texttt{AutoTVM} does not provide any way to constrain the search technique to using parameters that yield correct kernels.
We observe, for instance, that the \textbf{Grid} search often generates
invalid kernels, as the number of threads on the X, Y, and Z dimensions exceeds the total number of threads in the stream multiprocessor.

\begin{figure}[ht]
    \centering
    \includegraphics[width=0.9\textwidth]{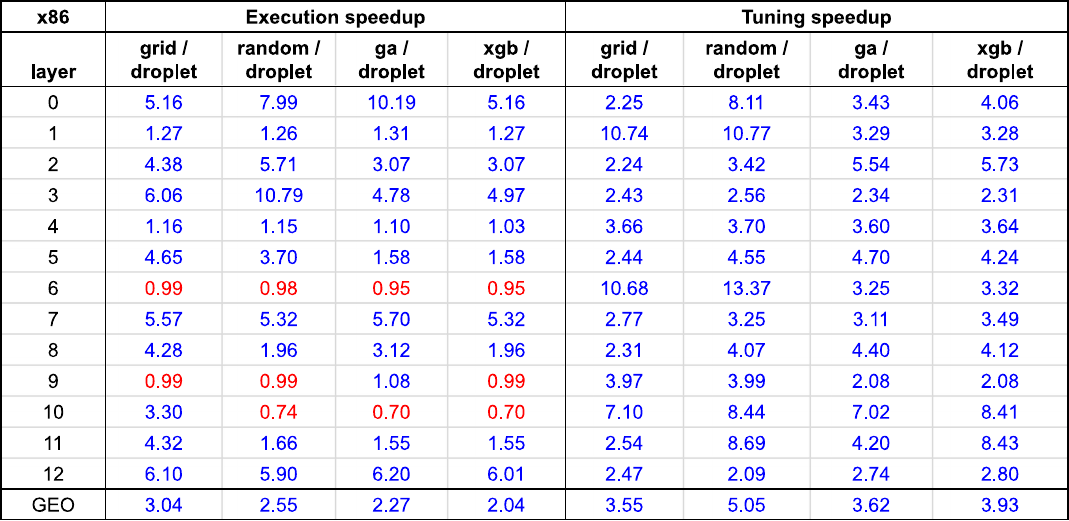}
    \caption{Results of Figure~\ref{fig:comparison_droplet_x86}, analyzed per layer of \texttt{AlexNet}.}
    \label{fig:layer_opt_x86}
\end{figure}

\paragraph{Discussion: Universality of these Results}
Droplet Search is not universally superior to the other search techniques available in \texttt{AutoTVM}, when used as an exploitation method in \texttt{Ansor}.
For instance, if we analyze the behavior of these different search approaches per layer of \texttt{AlexNet}, we observe situations where Droplet Search yields slower kernels than the other methods.
Figure \ref{fig:layer_opt_x86} demonstrates this point.
These subpar results happen due to the stochastic nature of all these autotuning techniques.
Droplet Search might stop on a suboptimal kernel because its convergence criterion is statistical in nature: if the running time of kernels is considered similar with a confidence level of 95\%, then the coordinate descent procedure stops.
However, regarding search speed, we have not identified one single layer of \texttt{AlexNet} where Droplet Search would be slower than the other search approaches.
Additionally, if we repeat the same analysis per layer of \texttt{AlexNet}
on the GPU setting, we observe that Droplet Search is superior to the other search approaches in every layer.
We omit these results from this paper for the sake of space.

\subsection{RQ5: The Average Number of Droplet Search Samples}
\label{sub:average_sample}

As hinted in Section~\ref{sub:search_technique}, Droplet Search has a convergence criterion discussed in Section 3.3 of its description~\cite{Canesche24}.
In this case, the search stops once there is no statistically significant difference between the current kernel and the kernels within its neighborhood.
Thus, the more optimized the seed of the coordinate descent algorithm, the fewer iterations it is intuitively expected to take until convergence.
This section investigates if this hypothesis is true.

\begin{figure}[ht]
    \centering
    \includegraphics[width=1\textwidth]{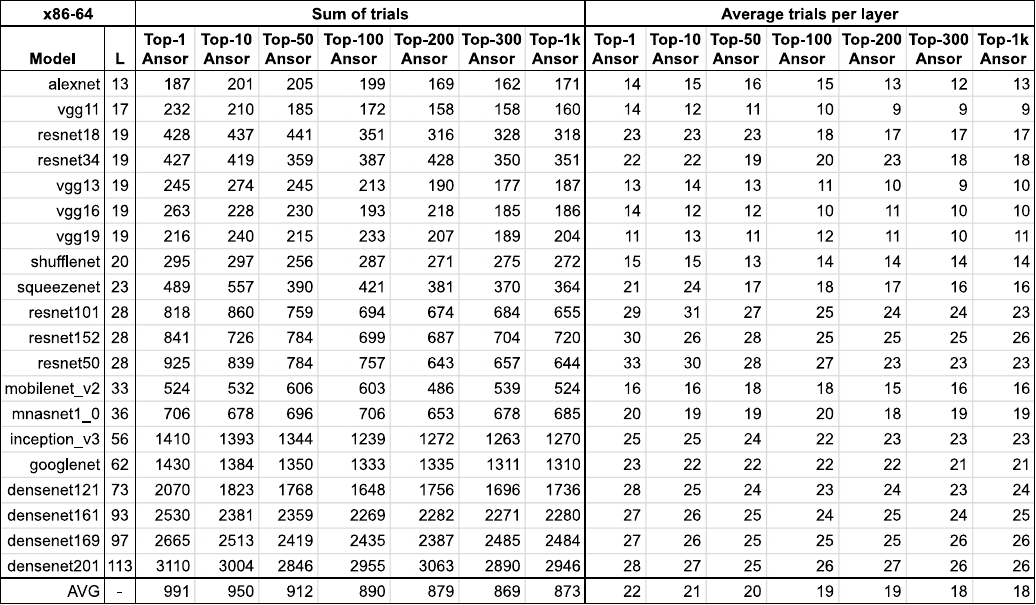}
    \caption{Number of trials sampled by Droplet Search until reaching convergence, considering the AMD x86-64 architecture.
    The average number of trials per layer divides the total number of trials per the number of layers in each deep learning model.}
    \label{fig:avg_trials}
\end{figure}

\paragraph{Discussion}
We count the number of iterations of Droplet Search on our different architectures, considering different budges for \texttt{DPAnsor}.
Figure \ref{fig:avg_trials} shows this number for each model evaluated on the x86 setting.
The figure reports a total number of trials and an average number of trials per layer.
The general tendency is that the larger the budget of trials allocated to
\texttt{DPAnsor}, the faster Droplet Search converges.

\begin{figure}[ht]
    \centering
    \includegraphics[width=1\textwidth]{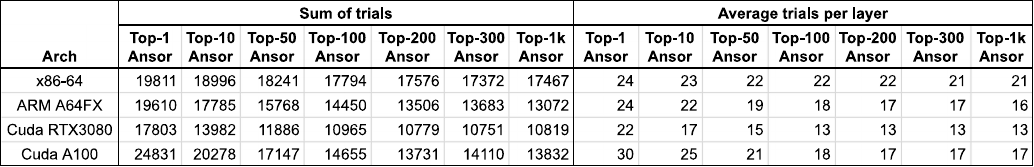}
    \caption{Average number of trials per model (considering 20 models)
    for different architectures.
    The averages per layer are the quotient of the total number of trials for all the models divided by the total number of layers.}
    \label{fig:avg_summary}
\end{figure}

Figure~\ref{fig:avg_summary} summarizes, for each architecture, the numbers earlier seen in Figure~\ref{fig:avg_trials}.
Figure~\ref{fig:avg_summary} reports averages per model (considering the 20 available models), and averages per layer.
In the latter case, we divide the total sum of trials observed for all the models by the sum of the number of layers present in every model.
In every case, the same tendency is evident: more trials sampled during exploration imply fewer trials sampled during exploitation.
This result is intuitive: as previously mentioned, Droplet Search tends to reach stability the closer to a local optimum it starts.

\subsection{RQ6: Comparison with Well-Known Machine-Learning Frameworks}
\label{sub:micro_kernels}

Thus far, the results we evaluated in this section are constrained to the \texttt{Apache TVM} software stack.
To provide the reader with some perspective on these results, we shall compare \texttt{Ansor} and \texttt{DPAnsor} with two well-known deep-learning frameworks: \texttt{PyTorch} 2.0~\cite{Paszke19} and \texttt{TensorFlow} v2.15.0~\cite{Singh20}.
Recently, \citet{Ansel24} have shown that \texttt{PyTorch} is able to outperform \texttt{Ansor} is many different workloads.
However, in \citeauthor{Ansel24}'s setting, \texttt{Ansor} was used as a backend, without autotuning the target kernels.
In this section we activate autotuning for \texttt{Ansor} and \texttt{DPAnsor}.
To this end, we give \texttt{Ansor} a budget of 1,000 trials.
In contrast, we use \texttt{DPAnsor}, either with a budget of 100 or 300 trials before activating Droplet Search.

\begin{figure}[hbt!]
    \centering
    \includegraphics[width=1\textwidth]{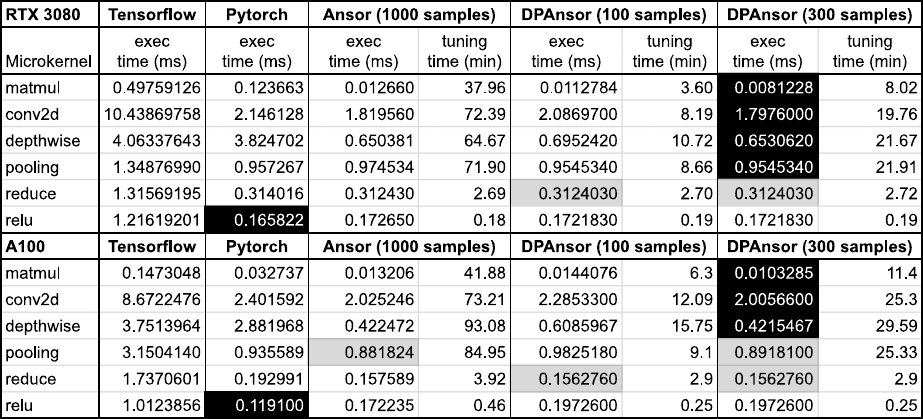}
    \caption{Comparison between the running time of kernels produced via \texttt{TensorFlow}, \texttt{PyTorch}, \texttt{Ansor} or \texttt{DPAnsor} running on two different GPUs.
    The black cells highlight fastest results;
    the gray cells indicate ties (with confidence level of 95\%).
    \texttt{TensorFlow} and \texttt{PyTorch} do not perform tuning.}
    \label{fig:micro}
\end{figure}

\paragraph{Discussion}
Figure~\ref{fig:micro} summarizes the results of the comparison of different deep-learning tools.
In our setting, \texttt{PyTorch} is consistently faster than \texttt{TensorFlow} on the two different graphics processing units available for these experiments.
However, \texttt{Ansor} and \texttt{DPAnsor} outperform \texttt{PyTorch} on large workloads (\texttt{matmul}, \texttt{conv2d}, \texttt{depthwise} and \texttt{pooling}).
Corroborating these results, the benefit of \texttt{Ansor} over \texttt{PyTorch} was also observed by \citet{Li21}.
Notice that these results do not include two kernels, \texttt{reduce} and \texttt{ReLU}.
In these two cases, neither \texttt{Ansor} nor \texttt{DPAnsor} have much room to perform optimizations.
These kernels show minimal memory reuse: they visit each memory cell only once, and little benefit can be acquired from autotuning.
As a consequence, the search time is very short when compared with the search time spent on the larger kernels.

\section{Related Work}
\label{sec:rw}

The design and implementation of systems to run machine learning models have been experiencing constant progress.
Following \citet{Ding23}, we recognize two main approaches to run deep-learning models, which we call {\it interpretation} and {\it compilation}.
In the former approach, frameworks such as \texttt{PyTorch}~\cite{Paszke19}, \texttt{TensorFlow}~\cite{Abadi16}, and the \texttt{ONNX} runtime~\cite{Jin20} implement machine learning models by binding operations to libraries such as \texttt{cuDNN}, \texttt{cuBLAS}, and \texttt{CUTLASS}.
In the latter approach, tools such as \texttt{TVM}, \texttt{Halide}~\cite{Kelley13}, \texttt{TorchInductor}~\cite{Ansel24} or \texttt{XLA} generate code for the operations that constitute the model.
This work concerns compilation; nevertheless, to provide some perspective to the reader about the relative effectiveness of these systems, Section~\ref{sub:micro_kernels} compares our approach with standard distributions of \texttt{PyTorch} and \texttt{TensorFlow}.
Contrary to the findings in Section~\ref{sub:micro_kernels}, \citeauthor{Ansel24} have recently shown that
\texttt{PyTorch}, in compilation mode, consistently outperforms \texttt{Ansor}.
However, in those experiments, \texttt{Ansor} was used solely as a code generator without performing any autotuning.
In our setup, once autotuning is enabled, \texttt{Ansor} can outperform \texttt{PyTorch} in most kernels.

\paragraph{Design and Construction of Autotuners}
The original presentations of \texttt{AutoTVM}~\cite{Chen18} and \texttt{Ansor}~\cite{Zheng20} outline the key techniques utilized in this paper.
These tools explore optimizations typically described via the {\it polyhedral model}~\cite{Feautrier96}.
Said optimizations, including fusion, tiling, and fission, are commonplace in tools such as \textsc{Polly}~\cite{Grosser12}, \textsc{Graphite}~\cite{Trifunovic10} and \textsc{PLuTo}~\cite{Bondhugula08}.
These tools do not solve the autotuning problem; however, they offer the basic infrastructure to do so, as \citet{Tavarageri21} have demonstrated with their \textsc{PolyDL} framework.

Recent research focuses on pruning the search space to enhance the efficiency of autotuners.
For instance, \citet{Tollenaere23}'s search algorithm uses a cost model to avoid exploring regions of the kernel space unlikely to yield good schedules.
However, building an accurate analytical model seems an illusive problem.
In the words of \citet{Ritter24}: ``{\it Modern processor designs use many techniques to improve overall performance that cause complex, irregular performance characteristics.}''
Additionally, there has been much work into adapting autotuners to deal with tensors whose shape is not statically
known~\cite{Mururu23,Pfaffe19}.
For instance, \textsc{DietCode}~\cite{Zheng22} and $\mathtt{SoD}^2$~\cite{Niu24} use a cost model similar to \citeauthor{Tollenaere23}'s, to represent the shape of a tensor as symbols that will only be known at runtime.
We notice that autotuning is not restricted to scheduling the order of computations within a kernel.
For instance, autotuning techniques can be used to choose the format to store tensors (sparse vs dense)~\cite{Ahrens22, Dhandhania21, Won23}, or the bitwidth used for quantization~\cite{Hubara21,Kloberdanz23}.

This paper does not propose a new kernel scheduling algorithm.
Rather, it is bringing forward the observation that the combination of two well-established heuristics tend to bring much benefit to the generation of high-quality kernels.
This benefit is measured not only in terms of the speed of the final code, but also in terms of the efficiency of the search technique.
In this sense, an important consequence of our work is the fact that it makes \texttt{Ansor} more hardware-aware.
By using coordinate descent as an exploitation approach, \texttt{Ansor}'s search remains circumscribed to a region that is likely to benefit more from the available hardware.
Some research groups have, independently, shown that the design of hardware-centric (in contrast to input-centric) autotuners tend to accelerate the search process~\cite{Zhu22,Ding23,Chendi24}.
This paper demonstrates this point without designing an algorithm that needs to be parameterized with hardware characteristics.

\section{Conclusion}
\label{sec:conclusion}

This paper has defended the thesis that state-of-the-art kernel scheduling methodologies can greatly benefit from a simple exploitation phase based on coordinate descent.
To support this thesis, we have implemented a combined exploration/exploitation search methodology in \texttt{Ansor}.
The new approach uses \texttt{Ansor} to explore different kernel optimization spaces and uses Droplet Search as a post-exploration phase.
This methodology improves \texttt{Ansor} and Droplet Search
in different ways:

\begin{itemize}
\item It enhances \texttt{Ansor}'s capability to exploit ``hardware boundaries''. The previous implementation of \texttt{Ansor} was not aware of the relationships between neighboring kernel schedules, as it lacked a concept of ``distance'' between the implementatios of kernels. The new exploration phase improves \texttt{Ansor}'s ability to better adjust optimization parameters to hardware constraints, such as cache sizes and vector widths.

\item It addresses Droplet Search's two limitations: its reliance on a well-defined {\it seed} (the initial kernel that initiates coordinate descent), and its inability to explore different kernel spaces. Previously, the seed and the kernel space were determined manually, requiring a programmer to provide Droplet Search with an initial annotated sketch~\cite{Canesche24,Chendi24}. The proposed methodology automates the seed generation process by utilizing \texttt{Ansor}.
\end{itemize}
As Section~\ref{sec:results} demonstrates, the proposed extension improves \texttt{Ansor} in terms of kernel quality and search time.
A container to reproduce those experiments is available at \url{https://github.com/lac-dcc/bennu}.
That implementation has been submitted to the \texttt{Ansor} community in late 2023~\cite{Canesche23}.
A similar extension was later deployed onto \texttt{TVM}'s \texttt{MetaSchedule}~\cite{Canesche24MS}, achieving even better results than  those seen in Section~\ref{sec:results}.
Thus, we believe that this combination of a wide exploration phase and a fine-grained exploitation step implemented via coordinate descent is general enough to be incorporated into different kernel schedulers.

\section*{Acknowledgment}
This project was sponsored by Cadence Design Systems. The authors extend their gratitude to Eric Stotzer and Vanderson Ros\'{a}rio for facilitating Cadence's financial support. Additionally, the authors acknowledge the support of CNPq (grants 314645/2020-9 and 406377/2018-9), FAPEMIG (grant PPM-00333-18), and CAPES (Edital \textsc{PrInt}).
The experiments in Section~\ref{sec:results} were conducted using hardware generously provided by the Ookami Computing Center, through US NSF grant \#192788, and by the Flatiron Institute, NY.
The authors express their appreciation to the TVM Community for the discussions that helped shape this work.

\bibliography{references}

\end{document}